\PassOptionsToPackage{table}{xcolor}

\documentclass[10pt,twocolumn,letterpaper]{article}

\usepackage{wacv}

\usepackage{multirow}
\usepackage{array}
\usepackage{tikz}
\usepackage{pgfplots}
\pgfplotsset{compat=1.18}
\usetikzlibrary{arrows.meta,positioning,fit,shapes,calc,backgrounds,shadows,patterns}

\DeclareMathOperator*{\argmax}{arg\,max}

\newcommand{\prov}[1]{#1}
\newcommand{\iexp}[1]{}
\newcommand{\starget}[1]{}

\newcommand{\TODO}[1]{}

\definecolor{bestbg}{gray}{0.84}
\definecolor{secondbg}{gray}{0.93}
\newcommand{\best}[1]{\cellcolor{bestbg}\textbf{#1}}

\newcommand{\fitw}{\ifdim\width>\linewidth\linewidth\else\width\fi}

\definecolor{wacvblue}{rgb}{0.21,0.49,0.74}
\usepackage[pagebackref,breaklinks,colorlinks,allcolors=wacvblue]{hyperref}

\def\wacvPaperID{2619}
\def\confName{WACV}
\def\confYear{2027}

\title{MotionSync: Non-Causal Refinement of Causal Tracker\\
for Label-Efficient 3D Perception}

\author{
Rahul Ahuja$^{*}$$^{1}$  \quad
Bala Murali Manoghar Sai Sudhakar$^{*}$$^{1}$  \quad
Shashwata Gupta$^{1}$ \\
Venkatraman Narayanan$^{1}$ \quad
Varun Ravi Kumar$^{1}$  \quad
Senthil Yogamani$^{1}$  \\[6pt]
{\tt\small $^{1}$Automated Driving, Qualcomm Technologies, Inc \quad $^{*}$Co-first authors} 
}

\begin{document}
\maketitle

\begin{abstract}
  Three-dimensional box-and-track annotation is the cost bottleneck in autonomous-driving data engines, and the offline systems built to relieve it replace the online perception stack outright, so a team needing both regimes maintains and reconciles two. \emph{MotionSync} makes the causal/non-causal boundary an explicit architectural seam instead. A strictly causal tracker, built on a strong published baseline~\cite{mctrack} and extended with innovation-driven uncertainty calibration, frame-rate-invariant kinematic association gates, and multi-hypothesis motion with learned mode selection, emits a valid online result. A non-causal pass then revises the buffered trajectories with Rauch--Tung--Striebel smoothing applied separately to pose, extent and yaw, physics-validated gap completion, and semantic pruning of ghost tracks against LiDAR point labels. The refiner never writes back, so one system serves both regimes and refinement's effect is a delta over an unaltered causal estimate. Used as an auto-labeller, a fixed 3D detector trained on \prov{25}\% human labels plus MotionSync pseudo-labels reaches \prov{96.9}\% of its full-supervision mean average precision (mAP) on Waymo, and at a \prov{10}\% budget the non-causal pass accounts for \prov{$+3.3$} mAP/L2 over pseudo-labels from the same tracker's causal stage. Re-fitting the online tracker on its own refined output recovers \prov{73}\% of the benefit of human supervision, while its causal output is worse supervision than no re-fitting at all. As a tracker MotionSync is at parity with the leading published offline entries on the headline metric and ahead of them on error composition, which is where a refinement pass can act at all: it reduces misses and fragmentations together, the signature of gap completion rather than of a tuned detector.
\end{abstract}

\section{Introduction}
\label{sec:intro}

A production autonomous-driving stack needs 3D perception twice, under two constraints.
On the vehicle it must be causal, seeing frames $1,\dots,k$ at frame $k$, a constraint
with a mature literature~\cite{ab3dmot,mctrack,fastpoly,li2023poly}. In the data engine
behind the vehicle it need not be: auto-labelling, simulation ground truth, benchmark
curation and trajectory mining run on complete logs, where causality is a self-imposed
handicap paid for in human annotation hours, the dominant marginal cost of scaling 3D
perception \cite{joseph2021autonomous, uricar2019challenges}.

The causal constraint has a specific and expensive failure. Under occlusion an online
tracker must commit, either terminating the track or coasting it forward, on evidence
predating the occlusion, and nothing lets it revisit that commitment when the object
re-emerges off the predicted path. For a vehicle that is an acceptable loss; for a label
store it is a fragmented trajectory a human must repair, at a cost set by the worst few
percent of tracks. Offline systems exploit the relaxed
constraint~\cite{huang2024bitrack,wu2023virtual,fan2023once,ma2023detzero} by replacing the
online pipeline outright, so a team needing both regimes maintains two systems and
reconciles their disagreements, and since the labels from one train the other, the
disagreement has consequences.

Our position is that the two regimes should share one tracker, differing only in whether
a refinement pass runs. \textbf{MotionSync} makes the causal/non-causal boundary an
explicit architectural seam. Stage one is strictly causal and valid on its own: we take a
strong published causal tracker, MCTrack~\cite{mctrack}, as the baseline and extend it
with uncertainty calibration, frame-rate-invariant association gates, multi-hypothesis
motion with learned mode selection, and motion-aware yaw estimation. Stage two is a
non-causal refinement block that revises the buffered trajectories with full temporal
context and never feeds back, so the causal estimate is final rather than revised.
The seam is what makes one system sufficient, and what lets the offline output supervise
the tracker that produced it; because the refiner is a strict post-process over a strong
baseline, its contribution is measurable as a delta over that baseline rather than as an
absolute benchmark position.

We contribute four things. \emph{The seam}: one tracker emitting causal and refined
outputs across a one-way boundary, so refinement's contribution is attributable by
construction (\cref{sec:method}). \emph{Frame-rate-invariant gates}: feasibility read
from the prediction residual in units of per-frame displacement, transferring from 10 to
2\,Hz (\cref{sec:association}). \emph{Per-attribute refinement}: Rauch--Tung--Striebel
smoothing on pose, extent and yaw separately, physics-validated interpolation, and
semantic pruning (\cref{sec:noncausal}). And \emph{a downstream evaluation}: detector
accuracy per unit of human annotation budget, and re-fitting the tracker on its own
refined output (\cref{sec:autolabel}). Of the four causal-stage components only the gate
parameterisation is claimed as a contribution; calibration, multi-hypothesis motion and
motion-aware yaw are engineering on a published baseline, and \cref{tab:ablation} reports
what each contributes so the refinement block's delta is not confounded with them.

Used as an auto-labeller on Waymo and nuScenes, a fixed detector trained on \prov{25}\%
human labels plus MotionSync pseudo-labels reaches \prov{96.9}\% of its full-supervision
mAP, and at a \prov{10}\% budget the non-causal pass alone accounts for \prov{$+3.3$}
mAP/L2 over pseudo-labels from the same tracker's causal stage (\cref{tab:labeleff}). As a
tracker it is competitive rather than leading: at parity on KITTI with the strongest
published offline entries, with the separation confined to the columns refinement acts on
(\cref{tab:summary}).

\section{Related Work}
\label{sec:related}

\paragraph{Causal 3D MOT.} AB3DMOT~\cite{ab3dmot} established the reference recipe, a 3D
Kalman filter with intersection-over-union (IoU) based greedy association, and later work
refined its halves: association measures degrading more gracefully under rotation and
depth error~\cite{mctrack,li2023poly, kumar2018near, fastpoly,kim2025grae3dmot}; noise models tied to
confidence or residuals~\cite{scorerefine,jiang2021adaptive}; richer dynamics and drift
mitigation~\cite{nagy2025motiondynamics,nagy2025robmot}; learned or hybrid filter
parameterisations~\cite{wang2026learntrack,dibella2025hybridtrack,pang2025pbmot,wei2026neural}; and
transformers replacing the filter outright~\cite{teye2025lidarmotdetr,teye2025futrtrack}.
All are strictly causal. Our causal stage adapts MCTrack~\cite{mctrack}, retaining its
category-specific structure and contributing not another association measure but a
frame-rate-invariant gate parameterisation and a calibration mechanism whose relation
to~\cite{scorerefine,jiang2021adaptive} we state in \cref{sec:calibration}. MCTrack also
reports an offline leaderboard entry (82.75 HOTA) above its online entry but does not
specify that mode's mechanism, so we take its published online entry as the causal
baseline and treat its offline entry as one of the leaderboard entries at or
above our own (\cref{sec:kitti}).

\paragraph{Offline tracking and offboard auto-labelling.} A smaller literature drops
causality: bidirectional association in BiTrack~\cite{huang2024bitrack}, our primary
tracking comparison; strong-detector pairings~\cite{wu2023virtual,wu2022casa}; identity
recovery across occlusion from future evidence~\cite{liu2023permanence}; multi-modal
correction passes~\cite{gu2024crosstracker, sobh2021adversarial}; annotation-first
pipelines~\cite{ding2026o3daa,li2025crowded}; and, closest in mechanism,
Offline-Poly~\cite{li2026offlinepoly}, which refines an upstream tracker within a
polyhedral framework and likewise examines recursive smoothing. Closest in \emph{purpose},
CTRL~\cite{fan2023once} and DetZero~\cite{ma2023detzero} are offboard \emph{detectors}
whose sequence-level refinement yields tracks as a by-product and whose stated use is
auto-labelling; they are not drop-in trackers, so they indicate what full-sequence learned
refinement can extract while remaining the relevant comparison for the application, the
structural difference being that they replace the causal stack whereas we post-process it.
Adjacent in purpose, unsupervised scene flow~\cite{ahuja2024optflow} extracts motion for
annotation without labels, per frame pair rather than per trajectory.

We therefore claim no novelty for bidirectional information, nor for applying the
Rauch--Tung--Striebel (RTS) smoother~\cite{rts} to tracking, which Offline-Poly studies
directly. Our contribution is narrower. The smoother runs \emph{per attribute}, pose,
extent and yaw each receiving their own recursion and strength because their error
characteristics differ (\cref{sec:rts}). It is paired with interpolation that tests whether
the filled motion is physically achievable, a test 2D gap filling~\cite{splineinterp}
cannot pose, the image plane admitting no dynamics to violate (\cref{sec:interp}). And the
refiner is a strict post-process, so the causal estimate survives intact, which is
what separates the seam from methods that replace the causal tracker. Non-causal operation
also admits a statistic no causal pass can form: a track's agreement with the semantic
labels of the points it encloses is measurable per frame, but the exposure-weighted mean
over its complete lifetime, on which the decision rests, is not (\cref{sec:semantic}).

\paragraph{Why a model-based refiner.} Learned sequence-level refinement is the alternative
and on Waymo it remains stronger (\cref{tab:summary}). We take the model-based route for
reasons specific to a data engine: it trains no sequence-level model, so there is no
sequence prior to transfer and no retraining when the detector changes; smoothing and
interpolation are $O(N)$ and parallel over tracks; and the causal estimate is left
untouched, which a learned end-to-end refiner cannot offer. The corresponding cost is
whatever such a model extracts beyond kinematics and semantics; the \prov{0.70}-point Waymo
gap of \cref{tab:summary} is a detector-confounded cross-system difference that loosely
bounds, rather than estimates, that quantity.

\section{Method}
\label{sec:method}

\begin{figure*}[t]
\centering
\includegraphics[width=0.78\textwidth]{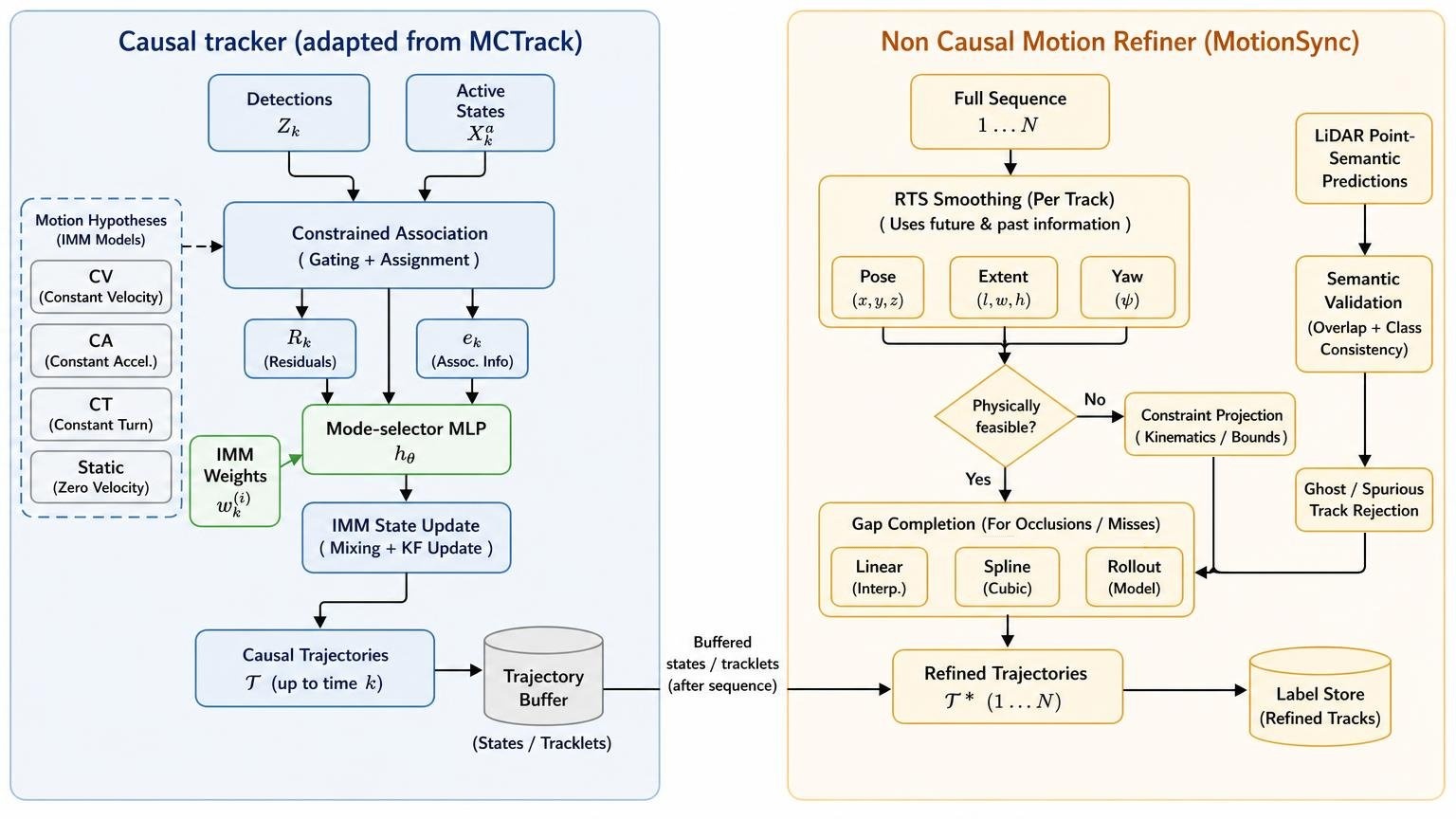}
\caption{\textbf{MotionSync architecture.} Stage~1 (blue) is strictly causal: predicted
active states and detections enter constrained association, whose residuals and
association statistics drive the mode-selector network $h_\theta$
(\cref{sec:calibration},~\cref{sec:imm}), and the causal trajectories $\mathcal{T}$ are
valid at every frame $k$ from $Z_{1:k}$ alone. They are also written to a buffer, which is the
seam: the boundary is crossed once, in one direction, after the sequence ends. Stage~2
(orange) consumes the buffered tracklets with the full sequence available, applying
Rauch--Tung--Striebel smoothing separately to pose, extent and yaw
(\cref{sec:rts}), testing whether a candidate fill is physically achievable before gap
completion (\cref{sec:interp}), and validating tracks against LiDAR point-semantic
predictions to reject ghosts (\cref{sec:semantic}). Nothing returns to Stage~1, so the
causal estimate is unaltered and refinement's effect on it is a pure delta; the
refined trajectories $\mathcal{T}^{*}$ are what a label store consumes.}
\label{fig:overview}
\end{figure*}

MotionSync pairs a causal tracker $\mathcal{C}$ with a non-causal refiner $\mathcal{R}$
(calligraphic throughout; boldface $\mathbf{R}$ later denotes measurement covariance).
Given detections $\{Z_k\}_{k=1}^{N}$, $\mathcal{C}$ produces $\mathcal{T}$ using only
$Z_{1:k}$ at frame $k$, and $\mathcal{R}$ produces
$\mathcal{T}^{*} = \mathcal{R}(\mathcal{T}, Z_{1:N})$ from the whole sequence, $\mathcal{T}$
carrying per-frame state, covariance and mode weights, all consumed by $\mathcal{R}$
(\cref{sec:rts}). Both are valid: $\mathcal{T}$ is what a vehicle consumes, $\mathcal{T}^{*}$ what a label store
consumes, and $\mathcal{R}$ never writes back (\cref{fig:overview}).

\subsection{Frame-Rate-Invariant Association}
\label{sec:association}

Under occlusion and in dense traffic the nearest detection is often the wrong one, and an
IoU cost cannot express that a match would require the object to reverse direction or triple
in size. Four terms supply that content, each parameterised so its threshold carries no
dependence on the sampling interval $\Delta t$, letting one gate set transfer between 10
and 2\,Hz (\cref{sec:nuscenes}).

A track carries a filtered bird's-eye-view (BEV) position
$\mathbf{p}_{k-1}\in\mathbb{R}^2$, velocity $\mathbf{v}_{k-1}$, extent $\mathbf{s}_{\text{track}}\in\mathbb{R}^3_{>0}$ and a
one-step prediction $\hat{\mathbf{p}}_{k|k-1}$; a detection carries
$\mathbf{p}_{\text{det}}$, $\mathbf{s}_{\text{det}}$. Two displacements must be
distinguished, and conflating them is the source of the frame-rate dependence removed here:
the \emph{implied displacement}
$\mathbf{u} = \mathbf{p}_{\text{det}} - \mathbf{p}_{k-1}$ grows linearly in $\Delta t$ by
construction, whereas the \emph{prediction residual}
$\mathbf{r} = \mathbf{p}_{\text{det}} - \hat{\mathbf{p}}_{k|k-1}$ does not, being the part
of the motion the prediction fails to explain, and it is what a feasibility test should
read. Below $v_{\min}$, $\mathbf{v}_{k-1}$ supplies no reference direction, so the
decomposition is dropped and bounds apply isotropically to $\|\mathbf{r}\|$.

\emph{Direction consistency} penalises
$c_{\text{dir}} = \lambda_{\text{dir}} \exp(\Delta\theta_v/\sigma_{\theta})$, where
$\Delta\theta_v = \arccos\!\big(\mathbf{v}_{k-1}^{\!\top}\mathbf{u} /
(\|\mathbf{v}_{k-1}\|\|\mathbf{u}\|)\big) \in [0,\pi]$; the exponential suppresses
near-reversals while leaving ordinary turning comparatively unpenalised, and read only
through an angle $\mathbf{u}$ contributes no dependence on scale or sampling interval. The
term is disabled when $\|\mathbf{v}_{k-1}\| < v_{\min}$ or
$\|\mathbf{u}\| < v_{\min}\Delta t$, where the angle is singular.

Two feasibility tests read $\mathbf{r}$. \emph{Lateral jump} treats the component of
$\mathbf{r}$ orthogonal to $\mathbf{v}_{k-1}$ as mis-association rather than manoeuvre,
rejecting outright beyond
$d_{\max} = \max(d_0,\, \gamma\,\|\mathbf{v}_{k-1}\|\Delta t)$: a floor that protects
parked vehicles plus a term proportional to the distance covered in one frame, so a single
$(d_0,\gamma)$ holds at any rate. \emph{Acceleration feasibility} penalises, again
exponentially, the implied constant acceleration above the per-category limit $a_{\max}$.
Under constant acceleration relative to a constant-velocity prediction
$\mathbf{r} = \tfrac{1}{2}\mathbf{a}\Delta t^{2}$, so the implied magnitude is
$2\|\mathbf{r}\|/\Delta t^{2}$ and the tabulated $a_{\max}$ are physical category limits
under that definition. Being in physical units $a_{\max}$ is invariant by construction,
though the finite-difference estimator behind it degrades as $\Delta t$ grows, which bounds
how much rate-invariance the parameterisation alone can deliver.

\emph{Extent consistency} exploits extent being near time-invariant, so a match
demanding a large size change is suspect:
\begin{equation}
c_{\text{size}} =
\begin{cases}
\infty & \text{if } \dfrac{\|\mathbf{s}_{\text{det}} - \mathbf{s}_{\text{track}}\|_{1}}{\max(\|\mathbf{s}_{\text{track}}\|_{1},\, s_{\varepsilon})} > \tau_{\text{size}}, \\[4pt]
\lambda_{\text{size}}\|\mathbf{s}_{\text{det}} - \mathbf{s}_{\text{track}}\|_2^2 & \text{otherwise,}
\end{cases}
\end{equation}
suppressing cross-category confusion, most commonly car/truck, which survives
category-partitioned matching because the partition is by \emph{detected} class. The floor
$s_{\varepsilon}$ is required: unfloored, a degenerate detection drives the ratio to
diverge and gates out every candidate. All four terms add to the baseline's Ro-GDIoU
cost~\cite{mctrack}, infinite entries removed before assignment so hard gates shrink the
problem; weights and thresholds are per category~\cite{li2023poly,fastpoly}, tabulated in
the supplementary with how often each gate fires.

\subsection{Online Uncertainty Calibration}
\label{sec:calibration}

Adapting $\mathbf{R}$ online is not new: it has been scaled by detector
confidence~\cite{scorerefine} and driven from observed residuals~\cite{jiang2021adaptive}.
Ours is narrower. Confidence-based scaling fits a map once and freezes it, and detector
confidence is not calibrated across weather, range or shift, a shortcoming recent detectors
address with uncertainty-aware fusion~\cite{narayanan2026mambafusion}; residual-based adaptation
estimates a covariance directly. We keep a learned confidence-conditioned \emph{prior}
plus one bounded scalar correction per attribute, stable on the short tracks that dominate
3D MOT.

An offline-fitted monotonic regressor supplies $\mathbf{R}_k^{(0)} = g(s_k, r_k, d_k)$
from confidence, range, and a depth/extent reliability proxy. Given innovation
$e_k = z_k - \hat{z}_{k|k-1} \in \mathbb{R}^{d}$ with predicted covariance $S_k$, we
hold exponential moving averages (decay $\lambda = \prov{0.9}$) $\mathrm{MAE}_k$ of
$\|e_k\|_2$ and $\hat{\sigma}_k$ of $\sqrt{\mathrm{tr}(S_k)}$, and rescale by their
ratio:
\begin{equation}
\mathbf{R}_k = \alpha_k \mathbf{R}_k^{(0)}, \quad
\alpha_k = \mathrm{clip}\!\left(\frac{1}{\kappa_d}\cdot\frac{\mathrm{MAE}_k}{\hat{\sigma}_k + \varepsilon},\ \alpha_{\min},\ \alpha_{\max}\right).
\end{equation}
The normalising constant $\kappa_d$ is what makes $\alpha_k = 1$ the fixed point of
correct calibration; omitting it leaves a perfectly calibrated 3D filter permanently
$5\%$ over-confident, the gate-too-tight direction, and its isotropy makes it an
\emph{upper} bound under the anisotropic covariances of range, where tracking is hardest.
Supplementary Sec.~B derives this, together with two consequences of applying a
standard-deviation ratio as a variance multiplier: the loop converges to $c^{1/3}$ under
true variance $c\,\mathbf{R}^{(0)}$, a deliberate damping that keeps it stable on short
tracks (\cref{tab:ablation} measures the squared variant), and the ratio carries an
$O(1/n_{\text{eff}})$ finite-sample bias dominated by the clip
$[\alpha_{\min},\alpha_{\max}] = [0.5, 3.0]$. So $\alpha_k$ is a bounded
diagnostic-driven correction, not a covariance-scale estimator; it is held at unity until
$n_{\text{warm}} = \prov{3}$ innovations accumulate. The update reads only past
innovations, so it runs online, and because it tightens
$S_k = H P_{k|k-1} H^{\top} + \mathbf{R}_k$ it sharpens Mahalanobis gating too, so
calibration and association compound. Scaling is per attribute, pose ($d=2$), extent
($d=3$), yaw ($d=1$), because their innovation statistics differ, not merely their
$\kappa_d$.

\subsection{Multi-Hypothesis Motion with Learned Mode Selection}
\label{sec:imm}

One motion model cannot describe a real trajectory: a vehicle accelerating from a stop,
turning, then parking exhibits three regimes, and a constant-velocity
filter~\cite{ab3dmot} is mis-specified in all but one. We maintain $M = 4$ hypotheses per
track, $\{\text{CV}, \text{CA}, \text{CT}, \text{Static}\}$, each a filter producing
$(\hat{x}^m_{k|k}, P^m_{k|k})$, and retain the interaction step of the classical
interacting multiple model (IMM): every mode is
re-initialised from the mixture of the others through the Markov transition matrix before
prediction, so an unused hypothesis cannot drift without bound. The four state spaces
differ (CT carries a yaw rate the others do not); the lifting is in the supplementary.

Classical IMM sets mode weights from that transition matrix and the hypothesis likelihoods
alone, ignoring two cheap signals: the recent innovation sequence indicates which model is
mis-specified, and scene context which manoeuvres are plausible. A lightweight network over
motion increments, previous innovation, confidence and a scene descriptor predicts
$\pi^m_k = h_\theta(\Delta\hat{x}_{k-1}, e_{k-1}, s_k, c_{\text{scene}})$, and the IMM
weights are reweighted and renormalised,
$\mu^m_k \propto \mu^m_{\text{IMM},k}\,\pi^m_k$, so coordinated-turn weight rises near
intersections without overriding the data; a uniform $\pi$ recovers classical IMM exactly
(architecture in the supplementary). Hypotheses are mixed, not selected, retaining the
outer-product spread term and the covariance inflation genuine regime ambiguity should
produce:
\begin{equation}
\begin{aligned}
\hat{x}_{k|k} &= \sum_{m} \mu^m_k \hat{x}^m_{k|k}, \\
P_{k|k} &= \sum_{m} \mu^m_k \Big(P^m_{k|k} + \delta^m_k (\delta^m_k)^{\!\top}\Big),
\end{aligned}
\end{equation}
with $\delta^m_k = \hat{x}^m_{k|k} - \hat{x}_{k|k}$. The module is causal and replaces
the single-model prediction step.

\subsubsection{Motion-Aware Yaw}
\label{sec:yaw}
Detector yaw error differs qualitatively between moving and static objects, so the cases
are treated separately, with circular statistics throughout because naive averaging is
wrong at the $\pm\pi$ wrap. For near-static objects yaw variation is almost entirely
detector noise and a circular median filter suffices. For moving objects direction of
travel is more reliable than box orientation, so yaw is estimated over a window of
$N_{\psi}$ frames (supplementary; $N_{\psi} \neq N$) as
$\hat{\theta}_{\text{motion}} = \arctan2(y_k - y_{k-N_{\psi}+1},\, x_k - x_{k-N_{\psi}+1})$,
its uncertainty taken from the circular variance of the frame-to-frame directions. Feeding
that variance into $R_\theta$ makes the estimate self-weighting, discounting erratic motion
at the static/moving boundary. The mechanism presupposes an observable yaw rate, which is
why articulated and slow classes remain difficult (\cref{sec:nuscenes}).

\subsection{Non-Causal Refinement}
\begin{figure}[t]
\centering
\includegraphics[width=\linewidth]{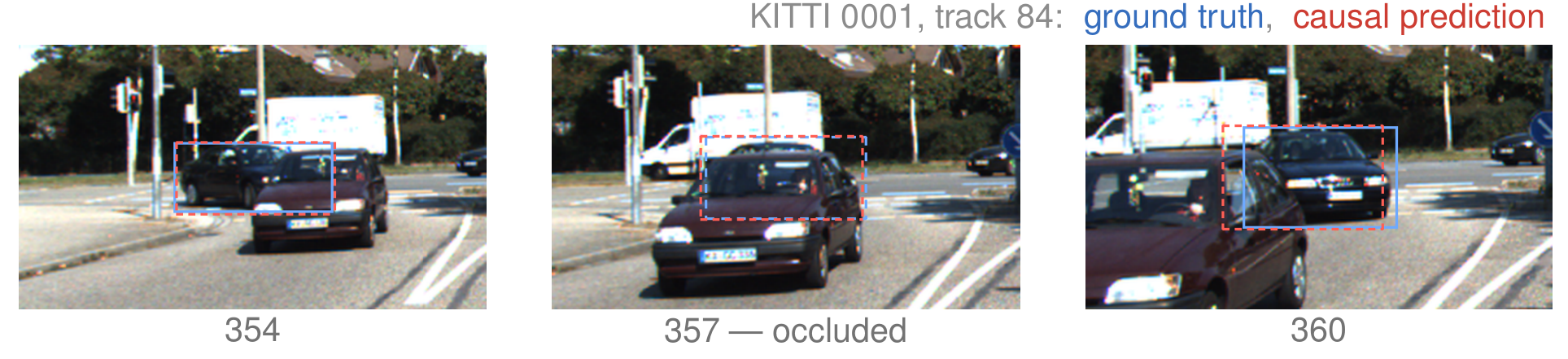}
\caption{\textbf{The failure the refiner repairs} (KITTI 0001, track 84). Entering
(354), during (357) and after (360) an occlusion, the causal estimate commits on
pre-occlusion evidence and drifts off the re-emerging vehicle (ground truth blue,
causal red); the non-causal pass revisits exactly this commitment.}
\label{fig:occlusion_real}
\end{figure}

\label{sec:noncausal}

\subsubsection{RTS Smoothing}
\label{sec:rts}
With the sequence complete, forward estimates are suboptimal by construction: each used
only a prefix of the data. The Rauch--Tung--Striebel smoother~\cite{rts} corrects
this in a backward recursion over mean and covariance, with gain
$\mathbf{A}_k = P_{k|k}\bar{F}_k^{\!\top}P_{k+1|k}^{-1}$:
\begin{equation}
\begin{aligned}
\hat{\mathbf{x}}_{k|N} &= \hat{\mathbf{x}}_{k|k} + \mathbf{A}_k\big(\hat{\mathbf{x}}_{k+1|N} - \hat{\mathbf{x}}_{k+1|k}\big), \\
P_{k|N} &= P_{k|k} + \mathbf{A}_k\big(P_{k+1|N} - P_{k+1|k}\big)\mathbf{A}_k^{\!\top},
\end{aligned}
\end{equation}
initialised at $k=N$ and run backward. We state the covariance recursion because
downstream stages consume the smoothed covariance, not the mean alone; $P_{k+1|k}^{-1}$ is
a ridged Cholesky solve, $P_{k+1|k}$ being near-singular for a track observed once. The
recursion is the smoother for one linear-Gaussian model while the forward pass emits a
moment-matched mixture, so we smooth that mixture with the mode-averaged transition
$\bar{F}_k = \sum_m \mu^m_k F^m_k$ (a GPB1-style approximation to the IMM smoother, not
the exact mode-conditioned recursion) and recompute mode weights from the smoothed
innovations. This is the refiner's one undischarged approximation, and its effect
propagates: a mode-averaged $\bar{F}_k$ attenuates genuine coordinated turns, which is why
the blend below is itself attenuated on turning tracks.

Applying the smoother to tracking is not new: Offline-Poly~\cite{li2026offlinepoly}
studies recursive smoothing for offline tracking directly. We run it separately on pose,
extent and yaw, at separate strengths, because their error characteristics differ: extent
is near-constant and tolerates aggressive smoothing, pose must retain genuine manoeuvres,
and yaw needs circular handling, unwrapped before smoothing and rewrapped to
$(-\pi,\pi]$ after, the recursion being defined on $\mathbb{R}$. Per-attribute smoothing
treats $P$ as block-diagonal across the three blocks, weakest exactly on coordinated
turns, where yaw rate and velocity direction are coupled by construction;
\cref{tab:ablation} measures it against a joint recursion. Estimates are blended as
$\mathbf{b}_k = (1-\beta)\hat{\mathbf{x}}_{k|k} + \beta\hat{\mathbf{x}}_{k|N}$, with
$\beta$ per attribute and reduced for pose on predominantly turning tracks. Smoother error
being orthogonal to any function of the data,
$\mathrm{Cov}(\mathbf{b}_k) = P_{k|N} + (1-\beta)^2(P_{k|k} - P_{k|N})$, exactly for the
linear-Gaussian case and approximately under the GPB1 step above. Downstream stages consume
that, not $P_{k|N}$, which would understate the uncertainty they gate on.

\subsubsection{Physics-Validated Interpolation}
\label{sec:interp}
Gaps are filled by a family selected on gap length (linear with circular yaw handling for
short gaps, clamped cubic spline for medium, kinematic rollout beyond), splines clamped at
both ends to the smoothed velocities of the bracketing observations so the segment joins
without a first-derivative discontinuity, and yaw on the unwrapped branch. Bracketing on
both sides is what offline operation makes available: implied accelerations are tested
against the $a_{\max}$ of \cref{sec:association} and a segment whose peak exceeds a fixed
multiple of it is discarded, on the principle that an unfilled gap is preferable to a
trajectory the object could not have followed; very long gaps are never filled. 2D gap
filling~\cite{splineinterp} cannot test this, no vehicle dynamics being violable in the
image plane. \Cref{tab:ablation} disables the test; the supplementary reports its firing
rate.

\subsubsection{Semantic Track Validation}
\label{sec:semantic}
Ghost tracks, persistent false positives on static structure, survive motion-based
filtering by being temporally consistent yet disagree with their enclosed point labels. With
$\phi_t^j$ the fraction of points inside track $\mathcal{T}_j$ of class $c_j$ labelled
$c_j$ at frame $t$, and $n_t$ the enclosed count,
\begin{equation}
\Phi^j = \frac{\sum_{t \in \mathcal{T}_j} w_t \phi_t^j}{\sum_{t \in \mathcal{T}_j} w_t},
\qquad
w_t = \min\!\left(n_t / n_{\text{ref}},\, 1\right),
\end{equation}
discounting sparsely populated frames; $\Phi^j_c$ replaces $\phi^j_t$ with the fraction of
enclosed points labelled $c$. Frames with $n_t = 0$ are excluded, and an entirely empty
track passes through unvalidated. The stage requires a minimum accumulated exposure
$\sum_t w_t \geq w_{\min}$ before acting, so one well-populated frame cannot delete a long
track: $\Phi^j$ is a weighted proportion, not an effective sample size. Normalising by
$\sum_t w_t$ rather than $|\mathcal{T}_j|$ is deliberate, the latter biasing $\Phi^j$
downward with the number of downweighted frames and so penalising distant tracks. Tracks
below threshold are rejected, or relabelled $c_j^{*} = \argmax_{c} \Phi_c^j$ when another
class explains the points, since deleting a car track sitting on a truck trades a false
positive for a false negative.

The labels are predictions, not annotations: a Cylinder3D~\cite{zhu2021cylindrical}
network trained on the \emph{training} split only
(SemanticKITTI~\cite{behley2019semantickitti, mohamed2020monocular} for KITTI, disjoint from the tracking test
split), so no evaluation-set label information enters the pipeline. This is the pipeline's
one dependency on labelled data, reported rather than resolved (\cref{sec:limitations});
where point-wise labels are unavailable the stage is disabled, at a cost
\cref{tab:ablation} measures.

\paragraph{Retrospective track management.}
\label{sec:trackmgmt}
Confirmation requires enough observations \emph{anywhere} in a lifetime, removing
artefacts an online tracker must tentatively emit, and trajectories on disjoint intervals
merge when the endpoint gap is short, one endpoint's extrapolation lands near the other,
and classes agree, recovering identities across occlusions no online gating window
tolerates. Nothing feeds back; settings in supplementary Sec.~C.

\section{Experiments}
\label{sec:experiments}

\subsection{Setup}
\label{sec:datasets}

KITTI~\cite{kitti} provides 21 training and 29 test sequences at 10\,Hz;
nuScenes~\cite{nuscenes} 1000 scenes split 700/150/150 at 2\,Hz, of which we evaluate the 7
tracked categories; Waymo~\cite{waymo} 798/202/150 sequences at 10\,Hz. We report
HOTA with CLEAR metrics on KITTI, average multi-object tracking accuracy (AMOTA) on
nuScenes and multi-object tracking accuracy (MOTA) on Waymo, alongside association accuracy
(AssA) and the CLEAR error counts: false negatives (FN), identity switches (IDS) and
fragmentations (FRAG). Every KITTI comparison value is the cited method's official
leaderboard entry (snapshot 2026-08-28); Waymo and nuScenes values are as published.

For auto-labelling the unit of analysis is the label set: MotionSync runs over the Waymo
and nuScenes training sequences, and the resulting tracks train one fixed 3D detector
(identical architecture, schedule and seed on every row) evaluated against human ground
truth on the held-out split. Waymo reports mAP/L2 and average precision weighted by heading
(APH/L2) over the three evaluated classes, nuScenes mAP and the nuScenes detection score
(NDS). The causal stage builds on MCTrack~\cite{mctrack} with its category-specific
parameters retained; $h_\theta$ is a two-layer multi-layer perceptron (MLP) trained on the
respective training split, and calibration clips $\alpha_k$ to $[0.5, 3.0]$. Gate thresholds and firing rates, refinement settings,
network details and the variance analysis are in the supplementary. Code, configuration files
and the pseudo-label generation scripts will be released.

\subsection{Auto-Labelling}
\label{sec:autolabel}

\begin{table*}[t]
\centering
\caption{\textbf{Label efficiency.} One fixed 3D detector (identical architecture,
schedule, seed) trained on human ground truth for a fraction of sequences plus
pseudo-labels for the rest, evaluated against human ground truth on the held-out split.
\emph{Row (c) is the decisive comparison}: same budget and detector as (b), differing only
in whether the refiner ran. Best per column shaded/\textbf{bold} among reduced-budget rows;
the 100\% row is the ceiling, not a competitor.}
\label{tab:labeleff}
\renewcommand{\arraystretch}{1.0}
\resizebox{\fitw}{!}{%
\begin{tabular}{llcccccc}
\toprule
\multirow{2}{*}{\textbf{Human labels}} & \multirow{2}{*}{\textbf{Pseudo-label source}} & \multicolumn{3}{c}{\textbf{Waymo}} & \multicolumn{3}{c}{\textbf{nuScenes}} \\
\cmidrule(lr){3-5}\cmidrule(lr){6-8}
 & & \textbf{mAP/L2}$\uparrow$ & \textbf{APH/L2}$\uparrow$ & \textbf{\% of ceiling} & \textbf{mAP}$\uparrow$ & \textbf{NDS}$\uparrow$ & \textbf{\% of ceiling} \\
\midrule
100\% & -- \emph{(ceiling)} & \prov{67.4} & \prov{64.8} & 100.0 & \prov{58.9} & \prov{66.2} & 100.0 \\
\midrule
50\% & MotionSync (full) & \best{\prov{66.6}} & \best{\prov{64.0}} & \prov{98.8} & \best{\prov{58.1}} & \best{\prov{65.5}} & \prov{98.6} \\
25\% & MotionSync (full) & \prov{65.3} & \prov{62.7} & \prov{96.9} & \prov{56.8} & \prov{64.3} & \prov{96.4} \\
\midrule
10\% & MotionSync (full) \emph{(b)} & \prov{62.4} & \prov{59.8} & \prov{92.6} & \prov{53.9} & \prov{61.6} & \prov{91.5} \\
10\% & causal stage only \emph{(c)} & \prov{59.1} & \prov{56.4} & \prov{87.7} & \prov{50.8} & \prov{58.7} & \prov{86.2} \\
10\% & none \emph{(d)} & \prov{53.6} & \prov{50.9} & \prov{79.5} & \prov{45.2} & \prov{53.8} & \prov{76.7} \\
\bottomrule
\end{tabular}%
}
\end{table*}

\subsubsection{Label Efficiency}
\label{sec:labeleff}
\Cref{tab:labeleff} is the paper's headline result, and its three-way structure is what
makes it interpretable: rows (b) and (c) train the same detector on refined and on
causal-stage pseudo-labels under an identical budget, schedule and seed, differing only in
whether the non-causal pass ran, while row (d) gives the reduced budget alone.

A quarter of the human budget recovers \prov{96.9}\% of full-supervision mAP/L2 on Waymo,
and nuScenes agrees; that figure, though, belongs to auto-labelling in general. What belongs
to this paper is the (b)--(c) gap at the tightest budget: at \prov{10}\% human
labels the non-causal pass alone accounts for \prov{$+3.3$} mAP/L2. Both rows run the same
tracker over the same logs, so the difference is attributable to a post-process that never
touched the online path. Were (b) and (c) to converge, the refiner would improve
nothing a downstream model uses and the FRAG and IDS reductions of \cref{tab:summary}
would be benchmark artefacts, so we report the comparison in the form that can fail. It
does not fail: the gap exceeds across-seed variation (\prov{$\pm0.3$} mAP/L2;
supplementary), and the refined-label rows close monotonically toward full supervision as
the budget grows. The reading we take is that pseudo-label \emph{quality}, not budget, was
the binding constraint on the detector, which is the claim that matters to an
auto-labelling user.

\subsubsection{Closing the Loop}
\label{sec:distill}
If the offline output can train a detector, it should also re-fit the tracker that produced
it. We re-fit the causal stage's learned parts (covariance prior $g$, mode selector
$h_\theta$, per-category gate thresholds) on four supervision sources drawn from sequences
\emph{disjoint} from the evaluation split, evaluating the resulting \emph{online} tracker
on the held-out split of \cref{tab:ablation} (full table in the supplementary). Refined
tracks recover \prov{73}\% of the benefit of human supervision.

The informative row is the third, and it is a sign rather than a magnitude: causal-stage
tracks are \emph{worse} supervision than not re-fitting at all. A module that decides which
motion model applies is actively misled by fragmented, identity-switching trajectories, so
the refinement pass is not merely additive. Without it the loop does not close, and that is
the strongest single result in the paper for the non-causal stage. The margins here are near
\prov{1} HOTA and we do not oversell them; the bootstrap interval is in the
supplementary.

\subsection{Tracking Quality}
\label{sec:sota}

Benchmark results establish that the tracker underneath the label generator is competitive;
they are not the contribution.

\begin{table}[t]
\centering
\caption{\textbf{Tracking quality: refinement as a delta over the causal baseline.} One row
per metric, each carrying its own direction, so no column mixes quantities.
MCTrack~\cite{mctrack} is the published causal tracker stage one extends, and $\Delta$ is
what refinement adds over that baseline on the same detections. The final columns name the
strongest \emph{published} entry we tabulate for that metric, so where we lead and where we
do not is visible directly. KITTI comparisons are official leaderboard entries (snapshot
2026-08-28), and the leaderboard carries further entries above our HOTA (\cref{sec:kitti});
nuScenes consumes identical CenterPoint~\cite{yin2021center} detections and its only
published comparison is the reproduced baseline itself; Waymo comparisons are as published,
where CTRL and DetZero are offboard detectors rather than drop-in trackers. The FN
comparison is restricted to the same-detector offline entries, because FN is not comparable
across detector rows: OC-SORT~\cite{cao2023observation} and
TripletTrack~\cite{marinello2022triplettrack} record 407 and 430 on 2D and monocular
detectors at a different operating point (\cref{sec:kitti}). The causal
stage's own KITTI figures are given in \cref{sec:kitti} and decomposed in
\cref{tab:ablation}; full per-class tables are in the supplementary.}
\label{tab:summary}
\renewcommand{\arraystretch}{1.05}
\setlength{\tabcolsep}{5pt}
\resizebox{\fitw}{!}{%
\begin{tabular}{llccrcl}
\toprule
 & & \textbf{MCTrack} & \textbf{MotionSync} & &
\multicolumn{2}{c}{\textbf{strongest published}} \\
\cmidrule(lr){6-7}
\textbf{Benchmark} & \textbf{Metric} & \textit{baseline} & \textbf{(ours)} &
\textbf{$\Delta$} & \textbf{value} & \textbf{entry} \\
\midrule
\multirow{6}{*}{\shortstack[l]{KITTI\\test (Car)}}
 & HOTA\,$\uparrow$    & 80.78 & \prov{82.94} & \prov{$+2.16$}  & 83.00 & Offline-Poly~\cite{li2026offlinepoly} \\
 & AssA\,$\uparrow$    & 84.30 & \prov{87.02} & \prov{$+2.72$}  & 86.39 & VirConvTrack~\cite{wu2023virtual} \\
 & MOTA\,$\uparrow$    & 89.82 & \prov{91.73} & \prov{$+1.91$}  & 93.19 & Offline-Poly~\cite{li2026offlinepoly} \\
 & FN\,$\downarrow$    & 1252  & \prov{578}   & \prov{$-674$}   & 702   & VirConvTrack~\cite{wu2023virtual} \\
 & IDS\,$\downarrow$   & 64    & \prov{13}    & \prov{$-51$}    & 8     & VirConvTrack~\cite{wu2023virtual} \\
 & FRAG\,$\downarrow$  & 438   & \prov{61}    & \prov{$-377$}   & 70    & Offline-Poly~\cite{li2026offlinepoly} \\
\midrule
\multirow{3}{*}{\shortstack[l]{nuScenes\\val (2\,Hz)}}
 & AMOTA\,$\uparrow$   & 74.0  & \prov{75.2}  & \prov{$+1.2$}   & --    & -- \\
 & AMOTP\,$\downarrow$ & 0.525 & \prov{0.498} & \prov{$-0.027$} & --    & -- \\
 & IDS\,$\downarrow$   & 275   & \prov{241}   & \prov{$-34$}    & --    & -- \\
\midrule
\multirow{3}{*}{\shortstack[l]{Waymo\\test}}
 & MOTA\,$\uparrow$    & 73.44 & \prov{74.35} & \prov{$+0.91$}  & 75.05 & DetZero~\cite{ma2023detzero} \\
 & MOTP\,$\downarrow$  & 22.78 & \prov{22.55} & \prov{$-0.23$}  & 22.24 & DetZero~\cite{ma2023detzero} \\
 & Miss\,$\downarrow$  & 18.04 & \prov{17.35} & \prov{$-0.69$}  & 17.17 & DetZero~\cite{ma2023detzero} \\
\bottomrule
\end{tabular}%
}
\end{table}

\subsubsection{KITTI}
\label{sec:kitti}
The pattern in \cref{tab:summary}, not the magnitudes, is the result. On the headline metric
we are at parity rather than ahead: the three closest published offline entries,
Offline-Poly~\cite{li2026offlinepoly} (to which we concede RTS-for-tracking novelty),
MCTrack's own offline entry and BiTrack~\cite{huang2024bitrack}, all lie within \prov{0.25}
HOTA of us, inside the \prov{$\pm0.42$} bootstrap interval of the supplementary, and the
leaderboard carries further entries above us. Parity is the honest reading and the expected
one: refinement cannot manufacture detections it was not given.

Where the system separates is error composition. Misses, fragmentations and identity switches
all fall steeply over the baseline, and reducing misses \emph{and} fragments together is the
signature of genuine gap completion: naive filling trades one against the other, so a method
improving both is recovering trajectory segments rather than relabelling boundaries. That is
what a label store consumes, and why \cref{sec:autolabel}, not this table, carries the
claim.

Three observations run the other way and bound the claim. Our MOTA trails
CasTrack~\cite{wu2022casa}: MOTA weights detection errors heavily, and pruning ghosts costs
precision on that metric even as it improves the label set.
VirConvTrack~\cite{wu2023virtual} reaches fewer identity switches by tracking
conservatively, paying 702 false negatives against our \prov{578}, a trade our operating
point declines to make. And fragmentation does not separate the causal stage at all:
LEGO's~\cite{zhang2023lego} 109 is far below our \prov{391}, so this result belongs to the
non-causal pass, not to the tracker underneath it.

\subsubsection{nuScenes at 2\,Hz}
\label{sec:nuscenes}
This comparison is detector-controlled, our tracker and the MCTrack baseline consuming
\emph{identical} CenterPoint~\cite{yin2021center} detections, and it isolates the one design
decision of \cref{sec:association} whose value is invisible on KITTI. With a lateral-jump
threshold in metres applied to the implied displacement, the pipeline lost 8.5 to 9.8 AMOTA
against that baseline. The mechanism is arithmetic: at 2\,Hz the same manoeuvre yields five
times the displacement, so a threshold fitted at 10\,Hz rejects valid matches. That the
loss was recall and not localisation is what made the diagnosis identifiable: AMOTP and IDS
were already best on this split under the old parameterisation. Reading the residual from
the prediction and bounding it in units of $\|\mathbf{v}\|\Delta t$ removes the dependence,
recovers the deficit and clears the baseline (\cref{tab:summary}). The lesson generalises: a
gate threshold expressed in metres is a hidden dependence on the sensor's frame rate, and
here that dependence cost more than everything the smoother contributed. Absent a controlled
decimation sweep the attribution remains an inference (\cref{sec:limitations}).

\subsubsection{Waymo}
\label{sec:waymo}
Waymo is where the refiner has most to work with, for a reason internal to the method: its
returns are the densest of the three benchmarks, so more points fall inside each candidate
box, and enclosed-point count is what semantic validation and the feasibility test consume.
Refinement improves every class over the baseline and cuts the miss ratio, the column gap
completion should move (\cref{tab:summary}). That the headline delta still reads smaller than
on KITTI is a property of the metric, not the sensor: MOTA counts mismatches but not
fragmentations, so it under-reports the error refinement principally removes, and the effect
surfaces in Miss instead.

We do not lead: DetZero~\cite{ma2023detzero} is first on six of seven columns and CTRL on the
seventh, leaving us second throughout. That comparison crosses a system boundary, their
margin reflecting a stronger detector as much as a different refinement strategy, so the
\prov{0.70} MOTA they hold bounds rather than estimates what learned sequence-level
refinement extracts beyond kinematics and semantics: a ceiling we sit close to while training
nothing.

\subsection{Component Ablation}
\label{sec:ablation}

\begin{table}[t]
\centering
\caption{\textbf{Component ablation}, all rows on one held-out subset of the 21 KITTI
training sequences (Car); no row is a test submission, so \emph{no cell is comparable with
\cref{tab:summary}}, only within-table differences. Components (\cref{sec:method}):
\textbf{C}~calibration, \textbf{A}~gates, \textbf{M}~multi-hypothesis, \textbf{Y}~yaw,
\textbf{R}~RTS, \textbf{I}~interpolation, \textbf{S}~semantic, \textbf{G}~retrospective.
Rows 1--5 build the causal stage, 6--9 add non-causal stages cumulatively; lettered rows
diagnose the row above; row 10 applies the refiner to an unmodified baseline.}
\label{tab:ablation}
\renewcommand{\arraystretch}{0.9}
\setlength{\tabcolsep}{4pt}
\resizebox{\fitw}{!}{%
\begin{tabular}{clcccc}
\toprule
\textbf{\#} & \textbf{Components} & \textbf{HOTA}$\uparrow$ & \textbf{AssA}$\uparrow$ & \textbf{IDS}$\downarrow$ & \textbf{FRAG}$\downarrow$ \\
\midrule
1  & \textit{baseline}          & \prov{79.6} & \prov{82.1} & \prov{121} & \prov{508} \\
2  & C                          & \prov{80.1} & \prov{82.9} & \prov{103} & \prov{494} \\
3  & C\,A                       & \prov{80.9} & \prov{84.0} & \prov{78}  & \prov{470} \\
4  & C\,A\,M                    & \prov{81.2} & \prov{84.5} & \prov{71}  & \prov{455} \\
5  & C\,A\,M\,Y                 & \prov{81.3} & \prov{84.7} & \prov{68}  & \prov{449} \\
\arrayrulecolor{black!30}\midrule\arrayrulecolor{black}
5b & \textit{as 5, classical IMM}          & \prov{81.1} & \prov{84.4} & \prov{73} & \prov{458} \\
5c & \textit{as 5, $\alpha$ squared}       & \prov{81.2} & \prov{84.6} & \prov{70} & \prov{452} \\
\arrayrulecolor{black!30}\midrule\arrayrulecolor{black}
6  & C\,A\,M\,Y\,+\,R           & \prov{82.0} & \prov{85.6} & \prov{44} & \prov{351} \\
6b & \textit{as 6, joint RTS}              & \prov{81.6} & \prov{85.0} & \prov{49} & \prov{372} \\
7  & C\,A\,M\,Y\,+\,R\,I        & \prov{82.6} & \prov{86.3} & \prov{31} & \prov{132} \\
7b & \textit{as 7, no feasibility test}    & \prov{82.4} & \prov{86.0} & \prov{36} & \prov{114} \\
8  & C\,A\,M\,Y\,+\,R\,I\,S     & \prov{82.9} & \prov{86.6} & \prov{27} & \prov{126} \\
9  & C\,A\,M\,Y\,+\,R\,I\,S\,G  & \prov{83.1} & \prov{86.9} & \prov{15} & \prov{68}  \\
\arrayrulecolor{black!30}\midrule\arrayrulecolor{black}
10 & \textit{refiner only:} R\,I\,S\,G     & \prov{81.4} & \prov{84.6} & \prov{62} & \prov{158} \\
\bottomrule
\end{tabular}%
}
\end{table}

\Cref{tab:ablation} decomposes the system on a single held-out subset of the KITTI
training sequences, every row on that split, so no difference crosses splits and no cell
matches the test figures of \cref{tab:summary}. Four diagnostic rows test the three claims
\cref{sec:related} narrows the contribution to. Gains here run larger than the
corresponding test-set deltas of \cref{sec:kitti}, as expected on a held-out subset of the
training distribution; only within-table differences are interpretable.

Row 6b substitutes one joint RTS recursion for the three per-attribute ones and is the
entire evidential basis for the per-attribute design; at zero the smoother should be joint
and simpler. Row 7b disables the feasibility test, and its \emph{shape} is the finding:
fragmentation improves as more gaps are filled, while HOTA and identity switches worsen
because some of those fills are wrong: the trade the design makes deliberately, a wrong
box being costlier for a human to repair than a missing one. Rows 5b and 5c bound the
learned parts: a classical IMM returns little, so $h_\theta$ is optional, and the squared
calibration ratio lands within noise of the damped form we ship (\cref{sec:calibration}).

Row 10 is the falsification row: the refiner over an unmodified baseline recovers much of
the full system but stays short of it, so the two halves compound rather than substitute.
Had it landed at or above row 9, the causal contributions would be redundant and the
two-stage framing wrong; we state that outcome in advance so the row cannot be read post hoc
as confirmation.
\vspace{-2mm}
\paragraph{Cost.}
\label{sec:cost}
Smoothing and interpolation are $O(N)$ and parallel over tracks, and the refiner trains no
sequence-level model, so label throughput scales with logged data rather than GPU budget.
The exception is semantic validation, whose per-frame segmentation pass dominates the
pipeline's cost; disabling it recovers most of that for the \prov{0.3} HOTA between rows 7
and 8 of \cref{tab:ablation}. Lacking matched end-to-end measurements for DetZero or CTRL we
make no quantitative cost comparison; the per-stage accounting is in the supplementary.

\section{Limitations}
\label{sec:limitations}

Four limitations warrant emphasis. First, refinement consumes a complete
sequence, so unbounded streams are out of scope and the seam offers the online path nothing
beyond leaving it untouched. Second, the frame-rate claim rests on the parameterisation argument of
\cref{sec:association} plus a before-and-after on one dataset rather than a controlled
decimation sweep, so the nuScenes attribution remains an inference. Third, semantic validation needs point-wise labels, hence a per-dataset segmentation
network: a system built to reduce annotation depends on one trained on annotations; it is disabled where such labels are unavailable, at
\prov{0.3} HOTA (\cref{tab:ablation}), and accounts for most of the pipeline's compute. Fourth, per-attribute smoothing treats the state
covariance as block-diagonal and the backward pass smooths a moment-matched mixture with a
mode-averaged transition, so both approximations are weakest on exactly the coordinated
turns the multi-hypothesis stage exists to model (\cref{sec:rts}).

\section{Conclusion}
\label{sec:conclusion}

MotionSync treats the causal/non-causal boundary as an architectural seam: a non-causal
refinement block sits on a strong published causal tracker and never writes back, so one
system serves both regimes where BiTrack~\cite{huang2024bitrack} and the offboard
detectors~\cite{fan2023once,ma2023detzero} replace the causal stack; one run yields
both the causal result and the labels that train the next detector. On novelty we are narrow: none is claimed for bidirectional information
nor for RTS~\cite{rts} smoothing in tracking, which
Offline-Poly~\cite{li2026offlinepoly} studies directly. Ours is the per-attribute
application, the physics-validated pairing, the frame-rate-invariant gates, and the refiner
remaining a strict post-process.

{
    \small
    \bibliographystyle{ieeenat_fullname}
    \bibliography{main}
}

\end{document}